\documentclass[preprint,3p,times,twocolumn]{elsarticle}

\usepackage{amssymb}
\usepackage{float}
\usepackage{booktabs}
\usepackage{multirow}
\usepackage{subcaption}
\usepackage[colorlinks=true]{hyperref}
\usepackage{xspace}

\usepackage{mathtools}
\DeclareMathOperator{\supp}{supp\,}

\makeatletter
\DeclareRobustCommand\onedot{\futurelet\@let@token\@onedot} 
\def\@onedot{\ifx\@let@token.\else.\null\fi\xspace}

\def\eg{\emph{e.g}\onedot} 
\def\ie{\emph{i.e}\onedot} 
 
 \def\vs{\emph{vs}\onedot}
 
\def\etal{\emph{et al}\onedot}
\makeatother

\journal{Neural Networks}

\begin{document}

\begin{frontmatter}



\title{CLAD: A realistic Continual Learning benchmark for Autonomous Driving}


\author[a]{Eli Verwimp\footnote{Corresponding author: \url{eli.verwimp@kuleuven.be}}}
\author[b]{Kuo Yang}
\author[b]{Sarah Parisot}
\author[b]{Lanqing Hong}
\author[b]{Steven McDonagh}
\author[b]{Eduardo Pérez-Pellitero}
\author[a]{Matthias De Lange}
\author[a]{Tinne Tuytelaars}

\affiliation[a]{organization={PSI, ESAT, KU Leuven}}
\affiliation[b]{organization={Huawei Noah's Ark Lab}}

\begin{abstract}
In this paper we describe the design and the ideas motivating a new Continual Learning benchmark for Autonomous Driving (CLAD), that focuses on the problems of object classification and object detection. The benchmark utilises SODA10M, a recently released large-scale dataset that concerns autonomous driving related problems. First, we review and discuss existing continual learning benchmarks, how they are related, and show that most are extreme cases of continual learning. To this end, we survey the benchmarks used in continual learning papers at three highly ranked computer vision conferences. Next, we introduce CLAD-C, an online classification benchmark realised through a chronological data stream that poses both class and domain incremental challenges; and CLAD-D, a domain incremental continual object detection benchmark. We examine the inherent difficulties and challenges posed by the benchmark, through a survey of the techniques and methods used by the top-3 participants in a CLAD-challenge workshop at ICCV 2021. We conclude with possible pathways to improve the current continual learning state of the art, and which directions we deem promising for future research. 
\end{abstract}




\end{frontmatter}


\section{Introduction}

Today, if a team of engineers were given the task to develop a new machine learning system for autonomous driving, they would start by defining the situations a car might encounter and should be able to handle. Data that fits the problem description is gathered and annotated, and a model is trained. Soon after deployment, reports of failure cases come in, \eg~the system fails in snowy weather, does not detect tricycles and misses passing cars on the right-hand side. In good faith, the engineers collect new data to include these situations, and retrain the model from scratch. Yet soon after the second version, reports come in of malfunctioning systems in cities for which no training data was collected, and the cycle starts again. Knowing how and where a model is going to be used, and how it will fail, is a nearly impossible task. While avoiding all possible failures is unattainable, decreasing the cost of incorporating new knowledge into a system is a much more feasible goal.		
\\ \\
When trying to develop such systems, the assumption is made that there exists a complete set of all the object-label pairs in the world, which includes those of interest. Observing these objects does not happen randomly, but depends on the context in which they are observed. For instance, the geo-location, the time of the day (\eg day \vs night) and the time period or era (\eg~1970’s~\vs now) influence the probability with which a certain type of vehicle is observed. Besides appearance, context also influences the frequency with which certain types of objects appear (\eg~fewer pedestrians on a highway). This is also true for machine learning datasets, where data is gathered within a context that ultimately determines which objects of the complete set are included in the machine learning process. 
\\ \\ 
Recognising that contexts are rarely constant, Continual Learning (CL) studies how we can enable neural networks to include new knowledge from changing contexts, at the lowest cost possible. Without continual learning, solving failure cases and extending to new domains requires retraining and tuning models from scratch. Given the increasing sizes of contemporary models, this has a significant energy and time cost~\cite{strubell2019energy}. A simple and straight-forward idea is optimising the model on the new data only. In 1989, McCloskey \etal~\cite{mccloskey1989catastrophic} were among the first that observed that this technique, finetuning, leads to rapid performance decreases on the old data. Today, this is referred to as catastrophic forgetting, and represents the largest challenge for continual learning: forgetting old knowledge cannot be the consequence of incorporating new knowledge in a system. 
\\ \\
With growing interest in continual learning, there is an increasing need for rigorous benchmarks and routes to reliably assess the progress that is being made. The earliest benchmarks focused on classification problems and split versions of popular datasets like MNIST~\cite{lecun1998mnist} and CIFAR10~\cite{krizhevsky2009learning}. In a typical \textit{split}-dataset, all available classes are artificially (and often randomly) divided into different tasks (or contexts). Then, a model is trained task-by-task, without access to past or future data. The performance of a CL-algorithm is assessed by the \emph{final} accuracy on each task. While they are useful for early methodological development, these benchmarks pose an artificial challenge. Randomly chosen distribution shifts are not assured to be aligned with real-world context shifts, which are the result of changing environments and model requirements. Additionally, datasets in image classification (\eg~CIFAR10) are often designed and simplified to only have a single object in the foreground, with a relatively small range of object scales. In Section~\ref{sec:cole_frame}, we describe how these benchmarks are only a small fraction of all possible CL-problems, and how other settings might be more realistic. 
\\ \\
In this paper we describe the design and the ideas motivating a new Continual Learning benchmark for Autonomous Driving (CLAD). CLAD-C tests a continual classification model on naturally occurring context shifts along the time dimension, and CLAD-D focuses on the more realistic problem of continual object detection. Both settings have been introduced as part of the ICCV SSLAD\footnote{\url{https://sslad2021.github.io/pages/challenge.html}} workshop in October 2021, for which we will present the top-3 submissions in this work. We start with a discussion on the principles we followed during the design of this challenge. We continue with a introduction of naive baselines, which highlight the difficulties and pitfalls of the proposed benchmarks. Then we discuss the solutions proposed by challenge participants and their results. Finally, we conclude with further details and experiments on the most promising submission ideas, discussion on future research directions to improve on the benchmarks, and continual learning in general.
Our contributions include: 

\begin{itemize}
    \item A review of current CL-benchmarks and their relations to one-another and shortcomings.
    \item Introduction of two new CL-benchmarks, with the goal of moving closer to CL-scenarios encountered in real-world problems. For both benchmarks code is made publicly available\footnote{\url{https://github.com/VerwimpEli/SSLAD_Track_3}}. 
    \item Review of the top performing methods on the proposed benchmarks, from the SSLAD ICCV '21 challenge entries, highlighting concrete pathways to progress the current state of continual learning.
\end{itemize}

\section{A Continual Learning Framework}
\label{sec:cole_frame}

Research interest in Continual Learning has increased substantially in recent years, resulting in the introduction of a growing number of benchmarks and desiderata. In this section we review the tasks and datasets that have been proposed and highlight those we deem most popular in terms of adoption at top tier Computer Vision focused conference venues.

\subsection{Continual Tasks}
\label{sec:rel_work:continual_tasks}
Starting from the complete set $S = \left\{X, Y\right\}$ of objects $X$ and their ground-truth labels $Y$, a machine learning dataset is defined as a subset $D \subseteq S$, and characterised by the probability distribution $P(X, Y | C)$. The context $C$ determines which object-label pair have non-zero probability of being in the dataset. In i.i.d.\,training, this context stays constant and therefore the probability with which samples are observed by the model also remain constant. In contrast, continual learning is characterised by a changing context $c$, which induces a shift in $P(X, Y | C=c)$. Often, each of these context changes is referred to as a \textit{task}. While in theory all kinds of context switches are a part of continual learning, only two types of context variations are commonly used to assess CL performance. In \textit{Class-incremental} learning, a new context causes $P(X)$ to shift, which induces a shift in $P(Y)$, and typically $\supp(Y_n) \cap \supp(Y_m) = \emptyset$ if context-ID $m \neq n$, meaning each class only occurs (has non-zero probability) during a single context. \textit{Domain-incremental} learning considers shifts in $P(X)$ that do not affect $P(Y)$, thus only the environment in which a class is observed changes. In some settings, the model has access to the context variable or context-ID during training and testing, which is referred to as \textit{task-incremental} learning. Finally, in this work we assume that $P(Y|X)$ is not subject to change, \ie the label $y$ of a sample $x$ never changes \cite{van2019three, lesort2021understanding}. See Figure~\ref{fig:cv_for_cl_benchmarks} for an overview and stratification of common benchmarks, including those proposed here.

\subsection{Contemporary CL for Computer Vision}
\label{sec:rel_work:cl_for_cv}

\begin{figure*}[t]
    \centering
    \includegraphics[width=0.8\linewidth]{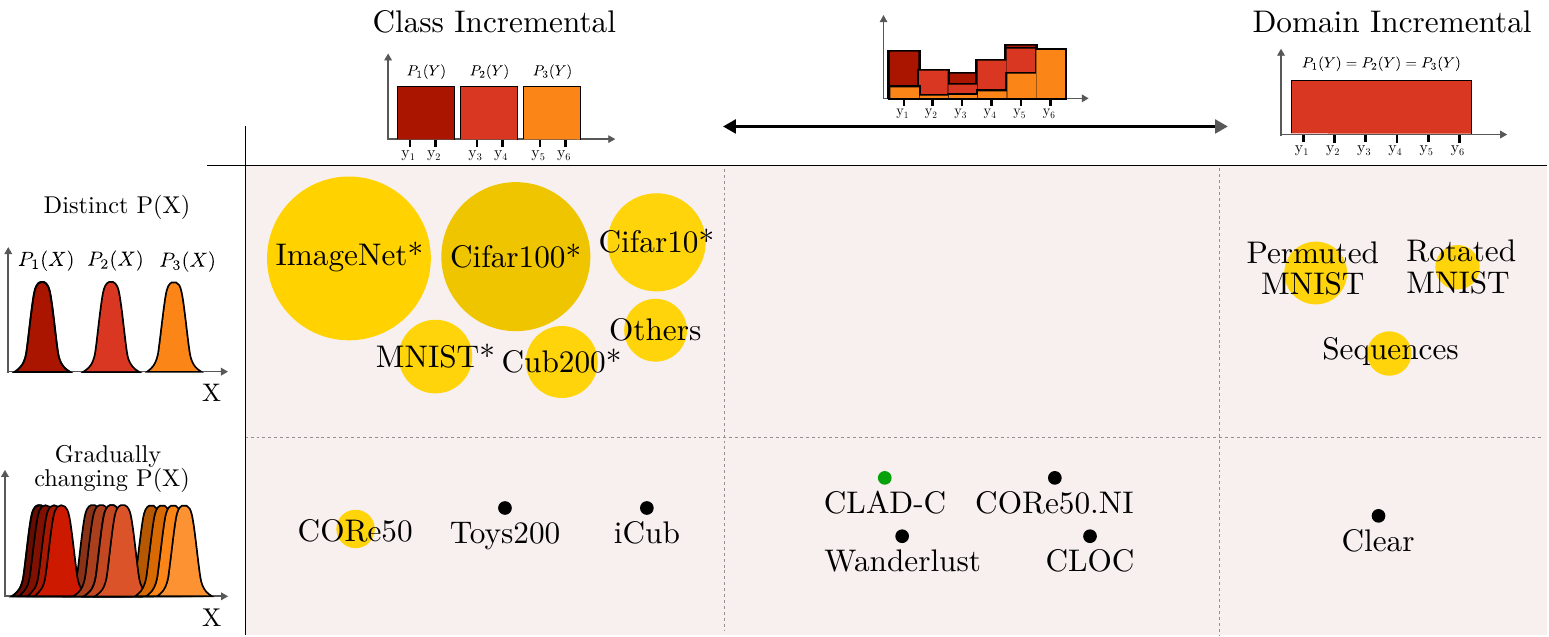}
    \caption{Overview of proposed and common benchmarks for continual learning in computer vision, with a focus on classification problems. We depict how data distributions $X$ and the available labels $Y$ change during training. The yellow circles areas are proportional to how many papers published in ICCV '21, CVPR '21 and NeurIPS '21 used each benchmark, see Section~\ref{sec:rel_work:cl_for_cv} for details. Black circles are recently proposed, more realistic benchmarks that are currently less popular. `*' indicates a `split' version of the dataset. `Others' refers to a set of benchmarks that were only used in a single paper and `sequences' refers to benchmarks where multiple datasets with the same labels are used. For more details and explanation of dataset placements, see Appendix.}
    \label{fig:cv_for_cl_benchmarks}
\end{figure*}

In theory, continual learning is concerned with any kind of distribution change in the training data. Given the many possible choices, many different benchmarks have been proposed and used. To understand which ones are currently popular in computer vision, we surveyed CVPR, ICCV and NeurIPS of $2021$; three highly ranked computer vision and machine learning conferences. From their proceedings, we selected all papers with the words \textit{continual, lifelong, sequential, incremental} or \textit{forget} in their titles. These keywords were defined using a manually collected list of $50$ papers concerning CL, spanning publication year $2017$-$2022$, of which 98\% had positive matches with our keywords. After filtering for false positives, $60$ relevant papers remain. 
Of these works, 73\% included at least one classification problem, 10\% a semantic segmentation problem, 7\% a generative problem, 3\% an object detection problem and 10\% various other problems. In the papers that focused on classification, $188$ experiments (excluding ablations etc.~) were conducted. Of these, 90\% used a non-continual dataset, and randomly change $P(X)$ at discrete intervals, such that each class has only non-zero probability in a single context (strictly class-incremental). 8.5\% changed $P(X)$ without affecting $P(Y)$ (strictly domain incremental), and only 1.5\% included more gradual context switches, see Section~\ref{sec:rel_work:towards_realistic}. See Figure~\ref{fig:cv_for_cl_benchmarks} for  the distribution of datasets used. While the random and discrete context switches in these benchmarks are only a small part of the space of CL-problems, they are currently used to assess the quality of almost all new CL-algorithms. 


\subsection{Towards more realistic benchmarks}
\label{sec:rel_work:towards_realistic}

While discrete and random distribution shifts are an interesting tool to study continual learning, they are, as previously discussed, not necessarily representative of the changing context in which continual learning systems can be used. In fact, as shown in Figure~\ref{fig:cv_for_cl_benchmarks}, there is a whole continuum of context changes between strict class and domain incremental learning. To aid progress towards applicable continual learning, Lomonaco \etal introduced CORe50~\cite{lomonaco2017core50}, and claim that realistic CL-benchmarks should have access to multiple views of the same object. This coincides with a more gradual changing context of the data distribution $P(X)$, an idea shared by~\cite{fanello2013icub, stojanov2019incremental,roady2020stream}, whom introduce iCub World, Toys-200 and Stream-51, respectively. Cossu \etal are critical of the lack of class repetition in continual learning benchmarks, and claim that this makes CL artificially difficult, and unlike real-world scenarios~\cite{cossu2021class}. They consider it realistic that a class has non-zero probability during multiple contexts, albeit with different frequencies. Additional support for this hypothesis can be found in related works~\cite{caccia2020online, lomonaco2017core50}. Repetition occurs naturally in benchmarks leveraging temporal meta-data of images, an idea implemented by the benchmarks Wanderlust~\cite{wang2021wanderlust}, Clear~\cite{lin2021clear} and CLOC~\cite{cai2021online}. Using time of day as a context variable, both the data and label distributions change gradually and non-randomly, which comes closest to a real-world setting, see Figure~\ref{fig:cv_for_cl_benchmarks} for how they compare to more traditional benchmarks. Regardless of context changes, some benchmarks only allow for online learning, where only a single pass over the data within a context is allowed~\cite{aljundi2019task}. This is regarded a realistic scenario in other works~\cite{De_Lange_2021_ICCV, cai2021online, yoon2021online, hayes2020lifelong}. Finally, using the context as an input variable, such that the model knows the context of a sample (task-incremental learning), has been critiqued as too restricting~\cite{aljundi2019task}. Despite these critiques and proposals to work towards more natural benchmarks, Section~\ref{sec:rel_work:cl_for_cv} indicated that they are seldom adopted in papers proposing new methods.
\\ \\
Most of these benchmarks focus on classification problems, as are most papers surveyed in Section \ref{sec:rel_work:cl_for_cv}. Despite its prevalence, classification is likely not the only scenario where CL will be applied in practice, since it often requires having a single (centered) object per image. Recent works \cite{peng2020faster,douillard2021plop} started exploring object detection and semantic segmentation in CL; two problems that are more likely to practically benefit from CL.

\subsection{Evaluation of Continual Learning}
\label{sec:rel_work:evaluation}

Besides good benchmarks, metrics that accurately reflect the goals of continual learning are indispensable. CL-methods are commonly evaluated using the average accuracy of each task at the end of training and average backward transfer (BWT); the difference between the accuracy of a task directly after it was trained and after all tasks were trained~\cite{lopez2017gradient}. These metrics are not necessarily aligned with the CL-goal of including new knowledge to an already working system. According to~\cite{lopez2017gradient}, the best algorithm is one that has the highest final accuracy, however we note that the BWT metric is also able to benefit from \emph{low} accuracy during training (\ie this effect improves the metric). Instead, we believe the best CL-algorithm is one that has high accuracy, which can be improved by including new data, without sudden performance drops. Ideally, this would be measured continuously during training, and the area below the accuracy curve used as a metric. Yet practically this has a high computational cost, and testing at well-chosen discrete intervals is a reasonable approximation. For further discussions, see~\cite{diaz2018don,kruszewski2020evaluating,caccia2020online}.

\section{Continual Learning Benchmark for Autonomous Driving}
\label{sec:benchmark}

In this section, we describe the design and development of our continual learning benchmark, using SODA10M~\cite{han2021soda10m}, an industry-scale dataset for autonomous driving. Self-driving vehicles can change urban mobility significantly, but there are still challenges to overcome. One such challenge is demonstrating the versatility of AI-based automated driving systems to cope with challenging and dynamic real-world scenarios. Several corporations are now driving many thousands of miles a day autonomously, creating streams of sensor measurements that form a natural source of continual learning data. This problem setting leads to natural and gradually changing distribution shifts, an excellent benchmark for CL-algorithms. Next, we introduce the SODA10M dataset, and provide a detailed overview of our two challenge benchmarks that utilise the available data.

\subsection{SODA10M Dataset}
\label{sec:benchmark:data}
Both tracks build on the SODA10M dataset~\cite{han2021soda10m}, which contains 10M unlabelled images and 20k labelled images. Image data consists of dash-camera recorded footage, obtained from vehicles driving through four Chinese cities, with images recorded at 10 second intervals. Ordering images chronologically largely entails visual footage of a car exploring the city and its neighbourhoods. The image label set has bounding box annotations for 6 object classes and covers different `domains' (cities, weather conditions, time of day and road type) -- see Figure \ref{fig:3a_collage}. See Figure \ref{fig:soda_10m} for some examples images (arranged in tasks for CLAD-D, as will be discussed in Sec.~\ref{sec:benchmark:trackB}). While self- and semi-supervised learning are interesting research directions~\cite{gallardo2021self}, we leave incorporating the unlabelled images for future work, and make use of the labelled dataset portion exclusively for our benchmark challenges.

\subsection{Challenge Subtracks}
\label{sec:benchmark:tracks}

As our goal involves working towards more realistic, real-world settings for continual learning, we approach the design of the challenge benchmarks with this mindset. As referred to in Section~\ref{sec:rel_work:continual_tasks}, there currently exist two main axes along which continual learning real-world task realism increases: firstly, the problem formulation itself and secondly, more realistic context shifts. Ideally, we combine these aspects into a single comprehensive benchmark. Yet, given that continual object detection is still in its infancy, we deem it more useful to make two separate benchmarks; along each axis, yet retain the potential to be combined in future. In the CLAD-C benchmark we firstly focus on naturally occurring distribution shifts using timestamps as a context variable, and in the CLAD-D benchmark we focus on continual object detection.

\subsection{CLAD-C: Online Continual Classification}
\label{sec:benchmark:trackA}

\begin{figure*}[t]
    \centering
    \includegraphics[width=.75\linewidth]{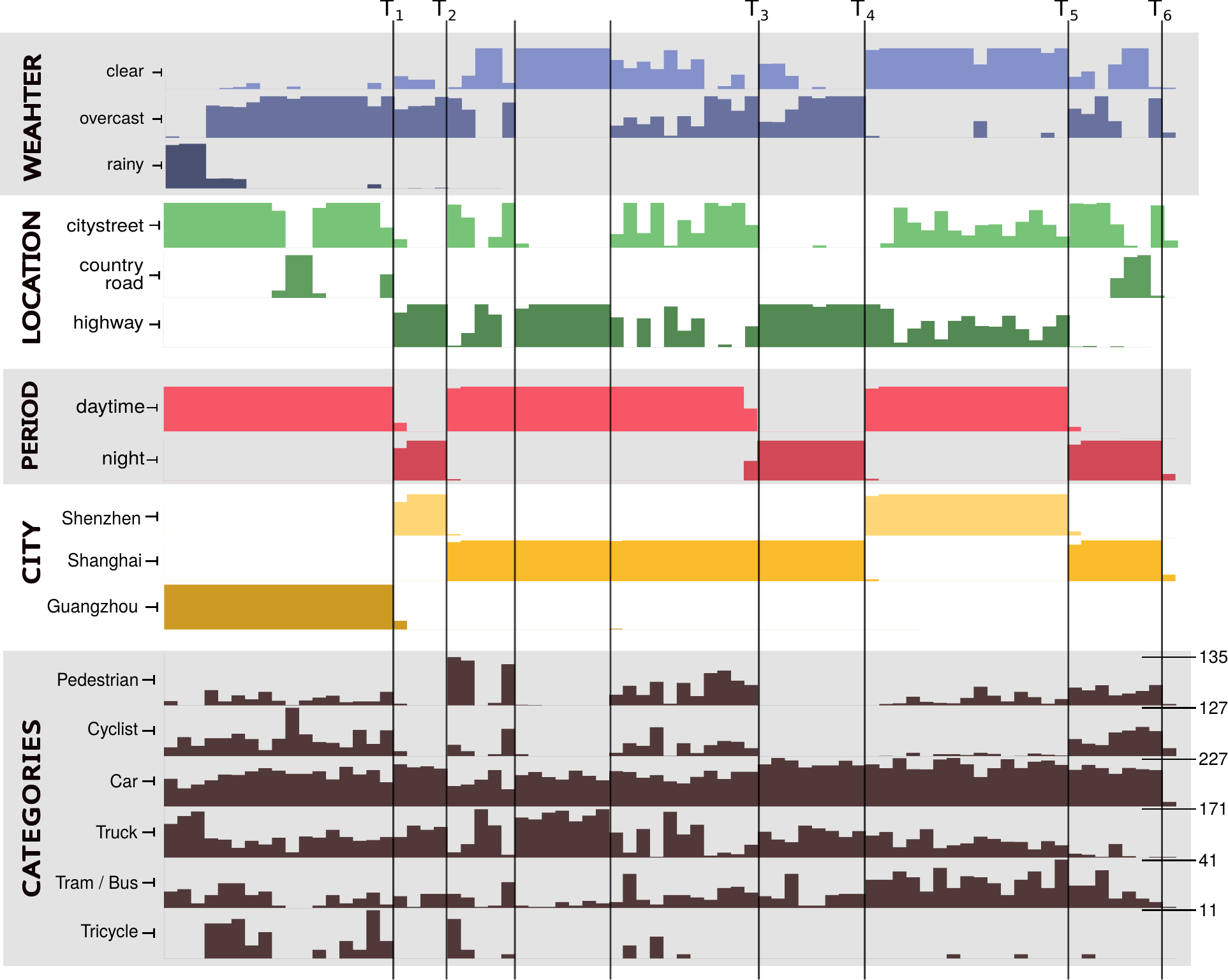}
    \caption{Overview of the domain shifts and the imbalance of the categories in the chronological CLAD-C benchmark. Objects are binned with 300 objects in each bin. The horizontal lines mark significant distribution shifts. The domain bars are all on the same scale, the scale of the categories is indicated on the right. }
    \label{fig:3a_collage}
\end{figure*}
 
\paragraph{Setup}
Following Aljundi \etal~\cite{aljundi2019task}, our first design principle covers the online setting, where explicit task boundaries are absent. Using the temporal meta-data of the images in SODA10M, we can construct an online stream of data spanning three days and nights. This stream exhibits domain shifts at different frequencies: changes from highway to city streets happen regularly and almost suddenly, while shifts from day to night are less frequent and more gradual. Finally the weather may change unpredictably and quickly. Gradually changing the data distribution causes gradual, but drastic, changes in the frequency with which different classes appear in the image stream (\eg at night and on the highway there are significantly few(er) pedestrians). The multi-view argument for benchmarks like CORe50, apply here as well. Since the images are at regular 10 second intervals, objects in the directly previous image and scene have a high likelihood to be present in the current image, likely from a different viewing angle. As Cossu \etal~\cite{cossu2021class} argue, repetition of past categories poses a realistic environment and scenario, which is reflected in the available chronological stream. Yet, aside from car objects, there are periods that all object classes disappear from the stream. At such moments, we conjecture that catastrophic forgetting can be prevented through the use of appropriate CL-strategies. See Figure~\ref{fig:3a_collage} for an overview of how the discussed domain phases switch and the corresponding categories most present during each phase.

To create the stream of object instances, we start from the chronologically ordered images and cut the target objects from their bounding boxes. We then change their aspect ratio to $1:1$ (padding along the shortest axis), add extra padding, and re-scale to $32{\times}32$. Occluded and too small ($<1024$ pixels) objects are removed. The final training stream consists of three alternating days and night domains, with a total of $22,249$ objects in $3841$ images. The test and validation set are from different days and nights, spanning all domains. The validation set contains $6305$ objects in $1590$ images, the test set has $69,881$ objects in $10,000$ images. For further details, see Appendix.

\paragraph{Evaluation}
\label{sec:benchmark:trackA:eval} 
As discussed in Section~\ref{sec:rel_work:evaluation}, we evaluate using the average accuracy of each class at given points during the stream. This also accounts for class imbalance; highly prevalent, ubiquitous classes (\eg cars) should not exhibit a higher weight in the final accuracy. Particularly, we test the model whenever there is a switch from day to night. If $A_{t, c}$ is the accuracy of class $c$ at time $t$, we define the Average Mean Class Accuracy (AMCA) as:

\begin{equation}
    AMCA = \frac{1}{|\mathcal{T}| C}\sum_{t \in \mathcal{T}} \sum_{c=1}^C A_{t, c}
\end{equation}
with $\mathcal{T}$ the set of all test times and $C$ the number of classes.

\begin{figure*}
\centering
\begin{subfigure}[t]{0.48\textwidth}
    \includegraphics[width=\textwidth]{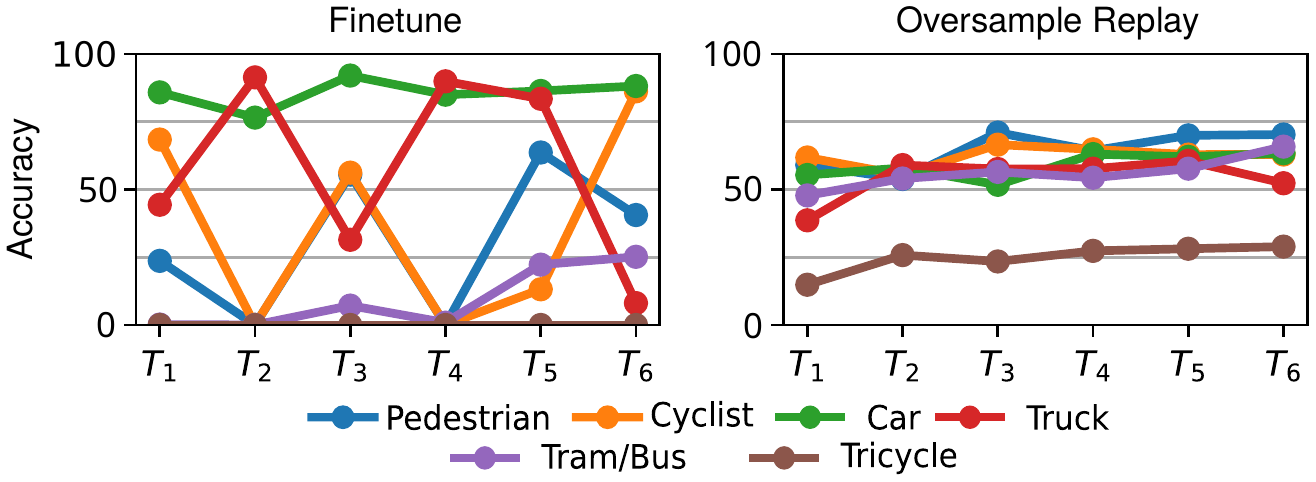}
    \caption{Test class accuracy for each class.}
\end{subfigure}
\hspace{1mm}
\begin{subfigure}[t]{0.48\textwidth}
    \includegraphics[width=\textwidth]{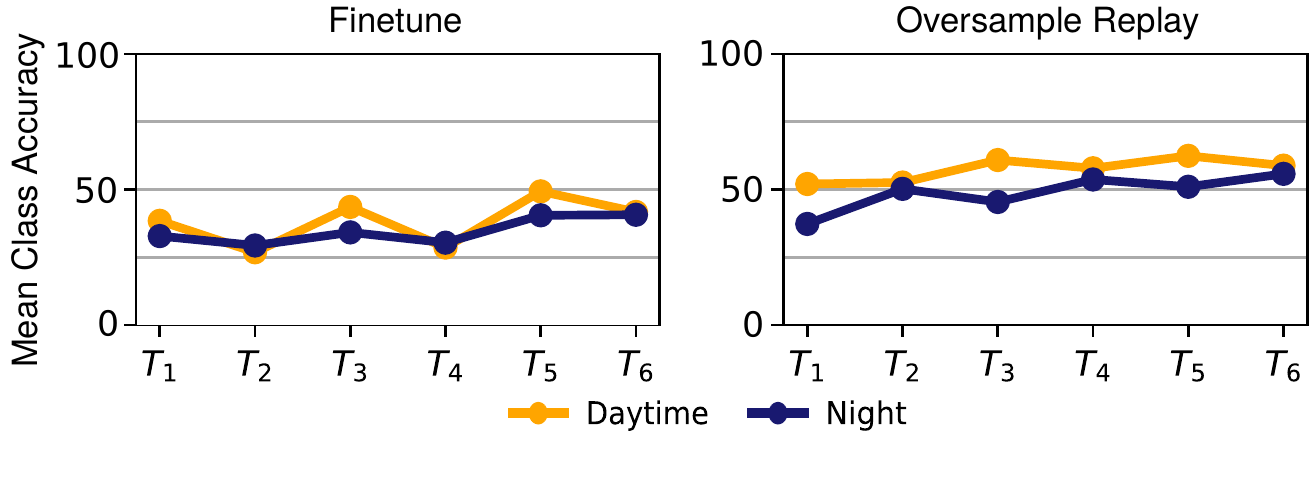}
    \caption{Test domain accuracy for day and night.}
\end{subfigure}
\caption{Class and domain accuracy for a simple finetune and an oversampling replay strategy on the CLAD-C at every day / night switch. The results of both baselines follow the data distribution closely, see Figure~\ref{fig:3a_collage}. The oversample strategy uses a memory buffer with $1000$ samples, and selects more samples of rare classes to balance the classes present in each batch.}
\label{fig:3a_results}
\end{figure*}

\paragraph{Baselines}
\label{sec:benchmark:trackA:baseline}
To assess the difficulty of the CLAD-C benchmark, we instantiate two baseline strategies. Naive finetuning is subject to catastrophic forgetting, where class accuracy's closely follow the class distribution present in the stream, as well as the shifts between night and day; see Fig.~\ref{fig:3a_results}. To overcome the large class imbalance, we implement an oversampling replay strategy. We allow $1000$ samples to be stored, divided equally among all classes. Before updating the model, each incoming batch is extended with samples from the memory such that the number of samples of each class is at least equal to $\frac{2b}{C}$, with $b$ the batch size and $C$ the number of classes. The finetune and oversampling benchmark reach respectively $36.8$ and $53.5$ AMCA. Both benchmarks were trained with an ImageNet pretrained Resnet50, using SGD and a batchsize of 10 (new samples).

\subsection{CLAD-D: Continual Object Detection}
\label{sec:benchmark:trackB}

\begin{figure*}[t]
\centering
\begin{subfigure}[t]{0.23\textwidth}
    \includegraphics[width=\textwidth]{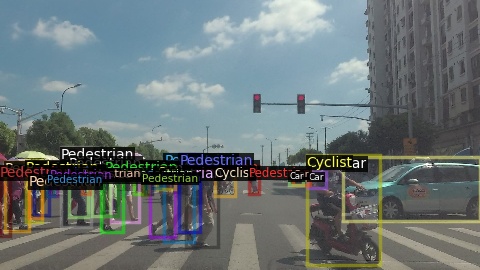}
    \caption{Task 1: Clear weather, at daytime, in the city center. $4470/500$ images.}
\end{subfigure}
\hspace{1mm}
\begin{subfigure}[t]{0.23\textwidth}
    \includegraphics[width=\textwidth]{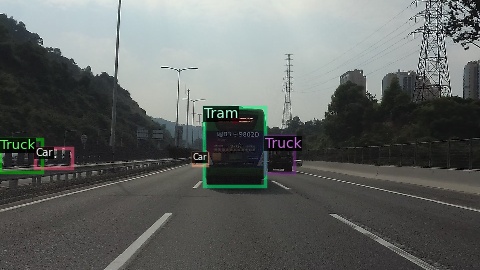}
    \caption{Task 2: Daytime images on the highway. $1329/148$ images.}
\end{subfigure}
\hspace{1mm}
\begin{subfigure}[t]{0.23\textwidth}
    \includegraphics[width=\textwidth]{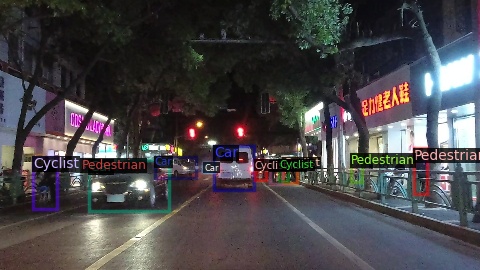}
    \caption{Task 3: Images at night. $1480/166$ images.}
\end{subfigure}
\hspace{1mm}
\begin{subfigure}[t]{0.23\textwidth}
    \includegraphics[width=\textwidth]{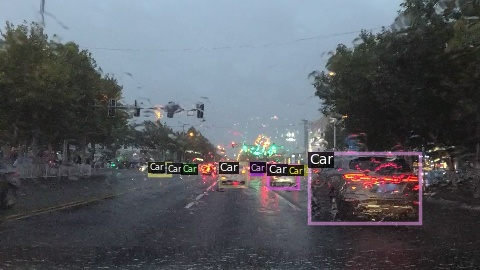}
    \caption{Task 4: Rainy images. $524/59$ images.}
\end{subfigure}
\caption{Example images and number of images in train/test set, for the four tasks in the domain incremental SODA10M benchmark. The displayed bounding boxes show the ground truth labels; colours indicate individual instances.}
\label{fig:soda_10m}
\end{figure*}

\paragraph{Setup}
\label{sec:benchmark:trackB:setup}
In Continual Object Detection (COD), class incremental benchmarks based on VOC and Microsoft COCO are common~\cite{peng2020faster, shmelkov2017incremental, joseph2020incremental}. An exception is Wang \etal~\cite{wang2021wanderlust}, whom introduce a benchmark similar to CLAD-C but for continual object detection, using first person videos. Given the dependence on large amounts of data in object detection, and the limited amount of labeled data in SODA10M, we designed a domain incremental benchmark for autonomous driving, which complements alternative pre-existing CL benchmarks. Our initial experiments (see Appendix) identify that \textit{highway}, \textit{night} and \textit{rainy} constitute domains that cause interference with the \textit{clear weather, daytime citycenter} domain, which has the most available data. We split the labeled data in four contexts: (1) clear weather, daytime on citystreets, (2) clear weather, daytime on the \textbf{highway}, (3) \textbf{night}, (4) daytime \textbf{rainy}. See Figure~\ref{fig:soda_10m} for canonical examples of each task. We emphasise that the model does not know to which domain or task an image belongs at test time, but we do allow multiple epochs on each task, making the problem more feasible.

\paragraph{Evaluation}
\label{sec:benchmark:trackB:eval}

We use the average mAP of each task at the end of training. Another option would involve measurement of accuracy after each task and take a mean, as in the CLAD-C benchmark, over all test times. We opted not to do this here, in order to both (1) retain alignment with current COD evaluation standards and (2) give each task equal weight. We report mAP at IOU level $0.5$, directly following the VOC benchmark~\cite{Everingham15}.

\begin{figure}[h]
    \centering
    \includegraphics[width=0.30\textwidth]{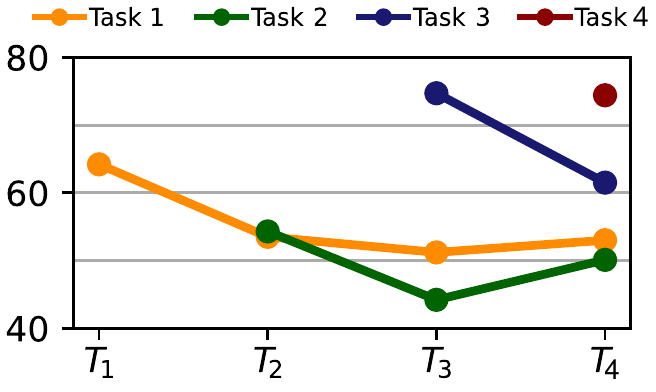}
    \caption{mAP for each task after training each task in the CLAD-D benchmark (\eg $T_1$ is after training the first task), without any continual learning strategy.}
    \label{fig:finetune_b} 
\end{figure}

\paragraph{Baselines}
\label{sec:benchmark:trackB:baseline}
We test the difficulty of the CLAD-D benchmark by finetuning each task on a FasterRCNN network, with an ImageNet Resnet50 backbone. The model is updated using SGD with a learning rate 0.01, momentum 0.9 and batch size 8. The learning rate is decayed by a factor 10 at epochs 8 and 11. The final task-average mAP is $59.8$. In Figure~\ref{fig:finetune_b} the result after training each task is shown, exposing task forgetting after learning. Although forgetting is not as catastrophic as typically observed in incremental classification settings, it may be observed that making effective use of all data in this setting poses a challenge.

\section{ICCV SSLAD challenge results}
Both the CLAD-C and CLAD-D were part of the ICCV SSLAD workshop in October 2021, for which respectively 49 and 55 teams participated. In this section we will summarise the techniques and ideas of the top three participants for each benchmark. For individual reports of each challenge, see the challenge's website.

\subsection{CLAD-C}

The top three submissions were G. Graffieti's team from the University of Bologna in Italy, A. Sabyrbayev of BTS Digital in Kazakhstan, and M.R. Kurniawan's team from Xi'an Jiaotong University in China. They respectively reached 68.6, 65.5, and 64.5 AMCA. All three have differences on multiple aspects, so a direct comparison of which techniques, choices or tricks made the difference is not possible. Instead, we make a qualitative analysis of their similarities and differences in this section. We refer to the top three submissions as C1, C2 and C3, see Table \ref{tab:a_results}. During the challenge some limitations were imposed: model sizes could not exceed the 105\% of the parameters of a ResNet50, pretraining was only allowed on Imagenet, batch sizes could be no larger than 10 new samples and any kind of exemplar memory is limited to 1000 samples. 

\begin{table}[]
\centering
\resizebox{0.4\textwidth}{!}{%
\begin{tabular}{@{}lccccc@{}}
\toprule
     & Fintune & Oversample & C1   & C2   & C3   \\ \midrule
AMCA & 36.8    & 53.5       & \textbf{68.6} & 65.5 & 64.5 \\ \bottomrule
\end{tabular}%
}
\caption{AMCA on CLAD-C for the baselines and top 3 participants}
\label{tab:a_results}
\end{table}

\paragraph{Architectures} Both C1 and C2 use a standard Resnet50 \cite{he2016deep}, while C3 uses a Resnet50-D. This was introduced in \cite{he2019bag} and improves ImageNet Top-1 classification with about 1\%-point, without adding more parameters. C2 also tested DenseNet169 \cite{huang2017densely} and EfficientNetV2 \cite{tan2021efficientnetv2}, but reported worse results on those networks, yet both perform better in non-sequential setups on CIFAR10 (which has the same image size). All models were pretrained on Imagenet. 

\paragraph{Optimizers} C1 and C3 use stochastic gradient descent (SGD) with learning rate $0.01$. C2 uses Adam \cite{kingma2014adam} with learning rate $0.0002$. C1 doesn't use momentum, weight decay or learning rate scheduling. C2 and C3 do not mention weight decay or momentum, but use multi-step learning rate schedulers. 

\paragraph{Data Augmentations} All three use horizontal flips. C1 adds upscaling to $224\times224$, which is the original Imagenet size. They report that this had a large influence on the final result, even though no information is added by doing this. They hypothesise this is due to the convolutional filters of the Imagenet-pretrained Resnet being calibrated to this size. C2 adds perspective transform and random erasing, and C3 adds random rotation and colour jitter. C2 also tested mixup \cite{zhang2017mixup}, but mention that this did not help. The original \textit{mixup} paper only reports a very modest improvement for a Resnet50 after 90 epochs, with larger improvements for larger networks and more iterations. 

\paragraph{Replay} All three winning submissions use some form of replay, but differ in the details of how the samples are stored and used. Replay is a very effective and efficient continual learning technique~\cite{chaudhry2019tiny}, but also the only one that was explicitly mentioned (and restricted) in the rules of the challenge. C1 reserves 100 samples per class in its memory, and selects samples using reservoir sampling \cite{vitter1985random} per class. C2 uses oversampling on the input stream, with manually set weights. This approximates reservoir sampling, but has fixed probabilities for samples entering/leaving the memory, while those of reservoir sampling are dynamically updated based on the number of samples seen of each class. C3 uses ideas from Rainbow Memory \cite{bang2021rainbow}. They calculate the entropy of each sample in a new batch, and then store 5 samples with the highest entropy. If the memory is full, the uncertainty for all samples is calculated under various augmentations. Then the 50\% most certain samples are removed. This deviates from \cite{bang2021rainbow}, where samples are stored across all levels of uncertainty. The idea is that this acts as a proxy to select diverse samples. C1, C2 and C3 use respectively 5+5, 4+6 and 10+6 new and old samples in each batch. 

\paragraph{Other Continual Learning techniques} Along with previously mentioned techniques, the three top submissions use other continual learning methods, briefly explained here. C1 employs CWR (Copy Weight with Reinit) \cite{maltoni2019continuous}, which stores the weights of the head after each task and restarts with randomly initialised ones. After each task, old and new weights are merged using an exponential moving average. This technically requires task boundaries, which were not meant to be used in the challenge. However, one could also use arbitrary intervals or detect task boundaries while training and then apply this technique. C2 uses a label smoothed version of the CE-loss, which has been hypothesised to create a better structured feature space \cite{muller2019does}. They have also tested CWR \cite{maltoni2019continuous} and LWF (Learning Without Forgetting) \cite{li2017learning}, but report worse results using both methods (while C1 does mention that CWR is important in their setup). C3 splits its training in two parts. The first part learns a metric space, using unsupervised and supervised contrastive losses \cite{chopra2005learning}. On top of this space, a classification head is trained using focal loss \cite{lin2017focal}, which is typically used in dense object detection to more easily learn hard classes (\eg classes with few samples). Together with the focal loss, they use a distillation loss on the past and current predicted logits for samples in the memory, which are, like the label smoothing in C2, temperature scaled. \\


\subsection{CLAD-D}
The top three participants were D. Li's team from the AML Group in the Hikivison Research Institute in China, M. Acharya's team from the Rochester Institute of Technology in the US, and J. Zhai's team from Nankai University in China. They respectively reached a task average of 74.7, 61.5, and 59.0 mAP, see Table~\ref{tab:b_results}. Hereafter, we will refer to these submissions as D1, D2 and D3. In CLAD-C the differences in the architectures used were relatively small, while their methods differed more. In contrast, D1 used a vastly different architecture than D2 and D3, while the method of D1 and D2 differ only in, nevertheless important, details. 

\begin{table}[]
\centering
\resizebox{0.28\textwidth}{!}{%
\begin{tabular}{@{}lcccc@{}}
\toprule
     & Fintune &  D1   & D2   & D3   \\ \midrule
mAP & 59.8    & \textbf{74.7} & 61.5 & 59.0 \\ \bottomrule
\end{tabular}%
}
\caption{Task averaged mAP on CLAD-D for the finetune baseline and top 3 participants}
\label{tab:b_results}
\end{table}

\paragraph{Architectures} D2 and D3 used the default \textit{torchvision} implementation of a FasterRCNN \cite{ren2016faster} model, with a Resnet50 feature pyramid backbone \cite{lin2017feature}, pre-trained on Microsoft COCO \cite{lin2014microsoft}. D1 changes the region proposal network to Cascade R-CNN \cite{cai2018cascade}, which has regression blocks trained at multiple IOU levels. With a Resnet101 backbone, this only gave marginal improvement. D1 then swapped the convolutional backbone to a transformer based one, a Swin Transformer \cite{liu2021swin}, which increased their results by more than 15 mAP points and almost completely alleviated forgetting (from 14.3 for Resnet101 to 1.7 mAP for the Swin Transformer). The baseline Swin Transformer has about 4 times the number of parameters than a Resnet50, yet the improvement is larger than expected from non-sequential results in \cite{liu2021swin}, where only a 6 mAP points gain was observed (on Microsoft COCO). 

\paragraph{Optimization and augmentation} D1 pretrained its backbone on ImageNet1K, with various augmentations, and then trained on the challenge's data for 50 epochs with a multistep learning rate scheduler. D2 and D3 use provided backbones from \textit{torchvision}, which are pretrained on Microsoft COCO. They also use multistep schedulers but use a cyclic schedule that restarts after each task.

\paragraph{Distillation} Both D1 and D2 use additional distillation losses \cite{hinton2015distilling} to transfer knowledge from a teacher network trained on a past task to a student network, trained on the new task. D2 only distills the output of the network, as in Shmelkov \etal~\cite{shmelkov2017incremental}. D1 is similar to Faster-ILOD \cite{peng2020faster} using distillation losses on the output of the backbone and RPN. Both use replayed images for distillation, while originally Faster-ILOD relies on the presence of unlabelled old samples in the new images.

\paragraph{Replay} Submission D1 stores 250 images and uses those for distilling the backbone and the RPN of the teacher. They do not mention how they select which samples they store. D2 stores for each task the images with the most pedestrians, cyclists and tricycles. They do not specify how they use replay exactly. D3 replays 2 old samples for each 2 new ones, and stores random images. They mention to also have tested balanced sampling, where they store images with the most tricycles, but this lead to worse results. 

\paragraph{Other} In D2, the outputs of the network are temperature scaled, which results in less confident predictions. According to the authors this results in better calibration and higher mAP.



\section{Discussion}

An important feature of a working continual learning solution will be the generality of the high-level features it can extract from an image. Simply optimizing a model to classify the available task data can lead a model down the garden path to a shortcut solution \cite{geirhos2020shortcut}. Such a solution only learns a, possibly meaningless, difference between the instances that are part of the current context. These differences might not suffice to discriminate current instances from future objects \cite{hendrycks2021natural}. Learning as many high-level concepts as possible from an instance, facilitates learning new instances, since the model has already learned concepts about past features that distinguish them from the new objects. If a first task consist of learning to classify lemons and limes, extracting the color of these fruits is probably sufficient. Yet, if the next task includes bananas, the model has to learn about the shape of lemons and bananas to tell the difference. This requires access to the past data which doesn't comply with the CL-setting. Among the techniques used by the participants we noticed that those that improve the generality of the extracted features had a large influence, which we'll discuss below. 
\\ \\
In CLAD-C all top three submission started from an Imagenet pretrained model, which has been shown to reduce forgetting in CL-systems due to its greater stability and better initial features \cite{mehta2021empirical, hu2021continual, wu2022class}. Even though they help, C1 shows that blindly using pretrained models can be suboptimal. Although the pretrained convolutional filters are scale invariant across a considerably large window, their performance degrades rapidly for objects below 100 pixels \cite{graziani2021scale}. Therefore it is useful to upscale the 32 by 32 objects of CLAD-C to the average 224 by 224 of ImageNet, not because it adds information, but because the model is tuned to those sizes (99\% of all ImageNet bounding boxes are larger than 100 by 100 pixels). 
\\ \\
The influence of having good general features might be most notably demonstrated by D1 on CLAD-D. Simply by employing a transformer based backbone, they show an impressive improvement. Intrinsically, vision transformers not only outperform convolutional based networks, they are also empirically more robust to changing input distributions and outliers, and less sensitive to high-frequency perturbations \cite{paul2021vision}. The potential for transformers in continual learning has recently been validated in several works \cite{mirzadeh2022architecture, yu2021improving, douillard2021dytox, pelosin2022towards}, but the slow convergence and need for large sets of \mbox{(pre-)training} data poses a serious challenge. 
\\ \\
Besides building upon pretrained models and better architectures, CL-systems should learn as much as possible from the available data, a limit that has not been reached in contemporary solutions \cite{prabhu2020gdumb}. C3 most explicitly tackles this challenge, optimising directly for a good feature embedding, on top of which a classifier is trained, similar to other works that use contrastive losses to build good representations \cite{cha2021co2l, pham2021dualnet}. Features learned this way have been shown to be more robust and better structured \cite{wang2020understanding, chen2020simple}, validated by the results in those works. Using a label smoothed loss, like C2 and C3, is also helpful in this regard. According to M\"uller \etal~\cite{muller2019does}, label smoothing creates a better conditioned features space: \textit{``label smoothing encourages the activations of the penultimate layer to be close to the template of the correct class and equally distant to the templates of the incorrect classes''}. An observation that is empirically validated in Dark ER \cite{buzzega2020dark}, who use a label smoothed version of the popular rehearsal strategy, which improves vastly over the standard approach. Finally, we note that the use of data augmentation, used by all top three participants, can decrease the number of shortcuts and lead to better representations \cite{buzzega2020dark, caccia2021reducing}.
\\ \\
Finally, we want to point out the prevalence of rehearsal in CLAD-C and distillation in CLAD-D. Rehearsal has been the dominant approach in continual classification, leading to many variants \cite{lopez2017gradient, chaudhry2019tiny, aljundi2019online, buzzega2020dark}, yet there is no clear answer to which samples are best in CL and how exactly they have to be used. GDUMP~ \cite{prabhu2020gdumb} even shows that it is a possibility the model is just re-learning to classify old classes by extracting all the information it needs from the memory. In detection, all methods rely on distillation of the model during training of new tasks. This is possibly influenced by SOTA methods like \cite{peng2020faster}, who focus largely on distillation. 

\section{Conclusion}
In this work we started with an evaluation of benchmarks that are currently used in Continual Learning, and how they relate to real-world scenarios where continual learning is most likely applied. We devised two benchmarks that increase the realism of CL-benchmarks, using less random and more gradual context shifts and by moving from classification to object detection. Both tracks were part of a challenge at ICCV '21, for which the top three participants gave us useful insights into the challenges and possible solutions for real-world CL applications. Finally, we discussed future directions of CL-research, including building better feature representations from the start and gaining a better understanding of CL's most important techniques, rehearsal and distillation, towards inspiring future improvement. We believe these to provide promising directions for future work. 


\bibliographystyle{elsarticle-num} 
\bibliography{references}

\newpage
\clearpage

\appendix

\section{Context and Continual Learning Benchmarks}
Here we will elaborate on how different changes in the data context lead to different kinds of continual learning, and how we conducted the survey on used benchmarks in continual learning benchmarks in ICCV, CVPR and NEURIPS '21, and a justification of the placements of the benchmarks in Figure \ref{fig:cv_for_cl_benchmarks}.

\subsection{Context as a framework for CL}
Assuming that the ground truth probability of observing a object-label pair $(x, y)$ is constant, the context $c$ in which they are observed can change the likelihood of observing samples. In i.i.d.\,training $c$ remains constant, and therefore $P(X, Y|C)$ as well. In continual learning, changing contexts are studied. In the two most used settings (class and domain incremental learning), these changes are artificially defined. In the former, each context only has non-zero probability for certain classes, and each class is only observed during a single context. For instance, split-CIFAR10 often starts with planes and cars, to be followed by deer and cats. In the latter, all classes are observed with the same frequency during each context, but their appearance changes. For instance, in MNIST-rotated, the probability of seeing each label does not change when the context change, but the images of the digits themselves are rotated. This means $P(X)$ has changed, but $P(Y)$ remained constant. Although both are interesting to study continual learning, they are only a small part of all possible context changes. More so, they are artificially designed, and therefore not necessarily the context changes that a model would encounter when it is applied on real-world problems. 
\\ \\
In contrast to artificially designed contexts, it is possible to use real-world contexts. For instance, in CLAD-C, by `observing' the available data chronologically, the context depends on the place where the picture was taken, the time of the day, and the weather. These variables have an influence on the appearance of the objects of interest, as well as the frequency with which they appear, see Figure \ref{fig:cv_for_cl_benchmarks}. Starting from non-random contexts, this blurs the boundaries between strict class incremental and domain incremental learning, as shown in Figure~\ref{fig:3a_collage}. Next, we describe how each of the benchmarks in Figure~\ref{fig:3a_collage} can be interpreted in the setting of changing contexts.

\subsection{CL-benchmarks and their contexts}
\begin{itemize}
    \item \textbf{Split-[dataset]}. The strict class-incremental split benchmarks in the top left corner of Figure~\ref{fig:3a_collage} all have similar changing contexts. Each class in the dataset only has a non-zero probability during a single context, and the context in which this happens is usually assigned randomly. The benchmarks do differ in the actual number of classes per context, \eg some papers use CIFAR10 with 5, 10 or 20 classes per context, while some include 50\% of all classes in the first context and then divide the others equally. 
    \item \textbf{Permuted and Rotated MNIST}. Both have been used since the first CL-papers (\eg~\cite{lopez2017gradient}) to assess domain incremental scenario's. They include all MNIST digits during each context, but respectively apply a pixel permutation or a rotation that changes, at discrete intervals (\ie tasks), with each new context. 
    \item \textbf{Sequences}. Sequences are sequences of datasets that include the same labels, but contain images from different domains. One example is the PACS benchmark~\cite{li2017deeper}, which has photos, art paintings, cartoons and sketches of the same classes, another is to use the shared classes in, for instance, ImageNet, CIFAR and VOC.
    \item \textbf{CORe50}~\cite{lomonaco2017core50} includes images from objects from different points of view. When these are observed by the model sequentially, this represents a gradual changing context: the appearance of the current object will first change slightly, before moving to the next object, which will again start gradually changing appearance. In the New Class (NC) scenario of CORe50, there is a broader context that determines which labels are included at what time. The New Instances (NI) scenario has every class in all contexts, but the surroundings in which the objects are observed change.  
    \item \textbf{Toys200 and iCUB-world}. Both benchmarks are similar to CORe50 new classes scenario. They include multiple views of the same object, which are observed sequentially. In Toys200 these are computer generated images from children's toys and in iCUB-world they are images from the point-of-view of a robot manipulating them.
    \item \textbf{Clear}~\cite{lin2021clear}. This benchmark is based on the YFCC100M~\cite{thomee2016yfcc100m} dataset, using objects that are subject to change during the period 2004 - 2007 (\eg handheld cameras changed quickly around that time). Each context only has images for a predefined set of objects from a single year between 2004 and 2007, \ie $P(Y)$ remains (approximately) constant, while $P(X)$ gradually changes.
    \item \textbf{CLOC}~\cite{cai2021online} is similar to CLEAR, and also based on YFCC100M. Like our CLAD-C, CLOC uses the timestamps of the images in YFCC100M to construct a stream of images. Rather than object classification, the goal here is to predict the geographic area the images was taken. 
    \item \textbf{Wanderlust}~\cite{wang2021wanderlust} uses images from an agent wearing a body came for several months, living their daily life. The images are sorted using their timestamps, which results in gradual changing $P(Y)$ and $P(X)$, depending on \eg the location of the agent and the time-of-day. 
\end{itemize}

\section{CLAD Details}
Here we provide further details on the distribution of objects in CLAD-C and CLAD-D.

\subsection{Split details}
Tables~\ref{tab:train_counts}, \ref{tab:val_counts} and \ref{tab:test_counts} contain the number of objects per class in each domain. For CLAD-C, the challenge doesn't evaluate per domain, so the more objects there are in a domain, the greater the influence on final AMCA score. Table~\ref{tab:day_counts} lists the date and period when the images used in the challenge were taken. This is primarily of interest for the training set, which is ordered chronologically. Finally, Table~\ref{tab:CLAD-D_object_counts} reports the number of objects in the train, validation and test set of the CLAD-D benchmark.

\subsection{Domain Gaps}
\begin{figure*}
    \centering
    \includegraphics[width=.65\textwidth]{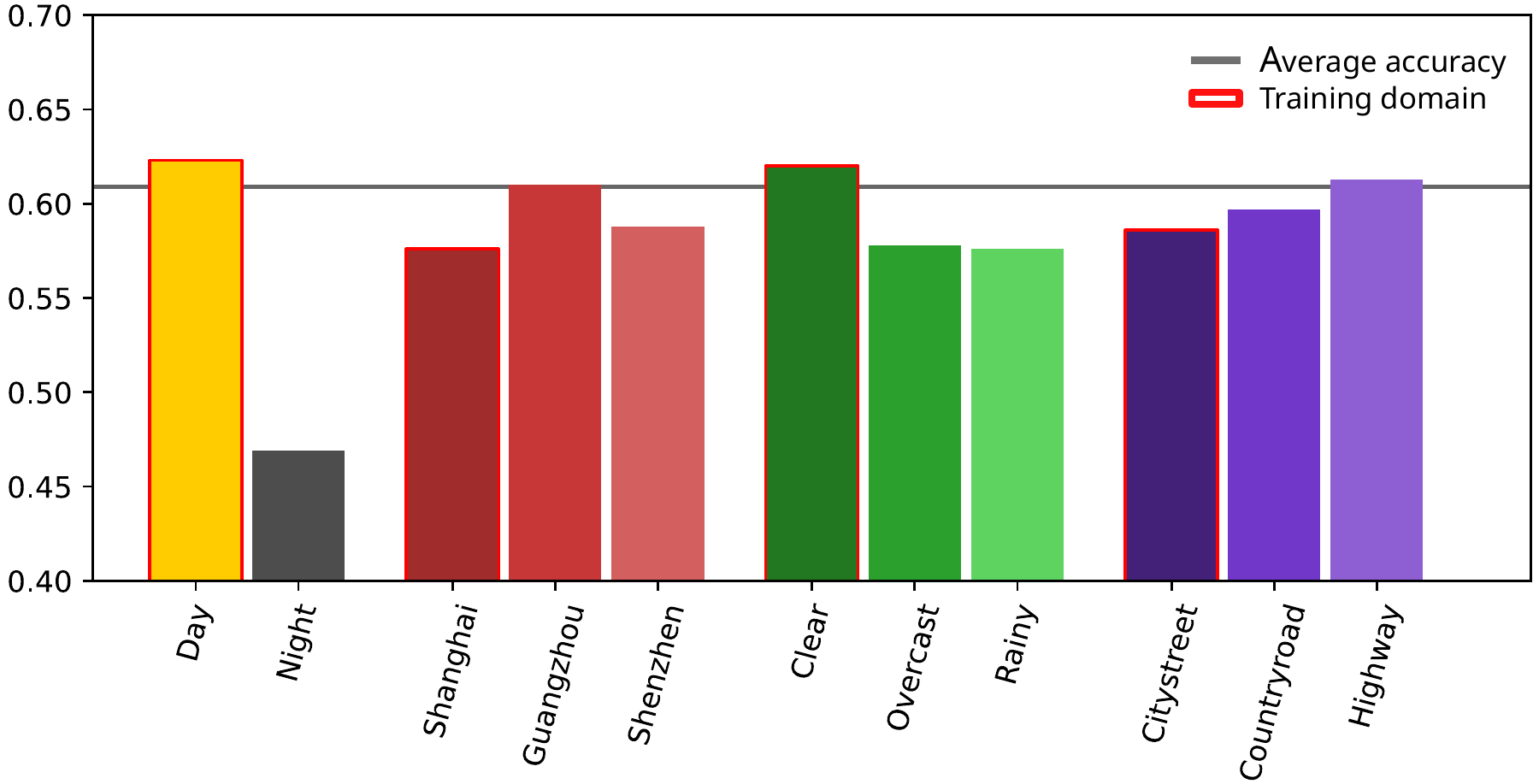}
    \caption{Average class accuracy of a Resnet50 trained only on images on Shanghai citystreet, daytime, clear weather images (indicated by the red stroke). The average accuracy is average only over the different classes and not the domains, \ie it is implicitly weighted by the number of objects in each domain.}
    \label{fig:domain_gaps}
\end{figure*}

Figure~\ref{fig:domain_gaps} shows the accuracy of a classification model that is trained on only clear weather, daytime imagery on the citystreets of Shanghai. The model used is a Resnet50, trained for 10 epochs on the available training data from this domain (see Table~\ref{tab:train_counts}). We used this initialy experiment to check where the largest domain gaps exist, which is clearly the night domain. These results do not mean that these are the only domain gaps, since transfer isn't symmetric. For instance, the transfer from the highway domain to the citycenter will have lower accuracy than the reverse, since there are less pedestrians and cyclists to train on in the highway domain.

\begin{table*}[t]
\resizebox{\textwidth}{!}{%
\begin{tabular}{@{}l|cc|ccc|ccc|ccc@{}}
\toprule
\textbf{train} & day  & night & Guanghzou & Shanghai & Shenzhen & clear & overcast & rainy & citystreet & countryroad & highway \\ \midrule
pedestrian     & 1933 & 462   & 360       & 1725     & 310      & 503   & 1878     & 14    & 2278       & 110         & 7       \\
cyclist        & 1530 & 300   & 1052      & 690      & 88       & 323   & 1176     & 331   & 1448       & 361         & 21      \\
car            & 7260 & 3028  & 2157      & 4845     & 3286     & 5517  & 4528     & 243   & 4541       & 902         & 4845    \\
truck          & 4691 & 1710  & 1341      & 3515     & 1545     & 3578  & 2323     & 500   & 1668       & 192         & 4541    \\
tram/bus       & 781  & 333   & 125       & 562      & 427      & 573   & 503      & 38    & 662        & 21          & 431     \\
tricycle       & 221  & 0     & 127       & 79       & 15       & 63    & 158      & 0     & 202        & 4           & 15      \\ \bottomrule
\end{tabular}%
}
\caption{Number of objects for each domain and each class in CLAD-C train set}
\label{tab:train_counts}
\end{table*}

\begin{table*}
\resizebox{\textwidth}{!}{%
\begin{tabular}{@{}l|cc|ccc|ccc|ccc@{}}
\toprule
\textbf{validation} & day  & night & Guanghzou & Shanghai & Shenzhen & clear & overcast & rainy & citystreet & countryroad & highway \\ \midrule
pedestrian          & 13   & 388   & 302       & 13       & 86       & 151   & 240      & 10    & 347        & 35          & 19      \\
cyclist             & 218  & 209   & 180       & 213      & 34       & 249   & 159      & 19    & 281        & 132         & 14      \\
car                 & 1329 & 1820  & 507       & 1315     & 1327     & 1266  & 1460     & 423   & 1153       & 787         & 1209    \\
truck               & 1542 & 117   & 32        & 1558     & 69       & 198   & 1219     & 242   & 72         & 155         & 1432    \\
tram/bus            & 86   & 371   & 247       & 86       & 124      & 133   & 288      & 36    & 201        & 176         & 80      \\
tricycle            & 200  & 12    & 12        & 200      & 0        & 212   & 0        & 0     & 212        & 0           & 0       \\ \bottomrule
\end{tabular}%
}
\caption{Number of objects for each domain and each class in CLAD-C validation set}
\label{tab:val_counts}
\end{table*}

\begin{table*}
\resizebox{\textwidth}{!}{%
\begin{tabular}{@{}l|cc|ccc|ccc|ccc@{}}
\toprule
\textbf{test} & day   & night & Guanghzou & Shanghai & Shenzhen & clear & overcast & rainy & citystreet & countryroad & highway \\ \midrule
pedestrian    & 2028  & 3262  & 2178      & 1034     & 2078     & 3084  & 1581     & 625   & 4553       & 217         & 520     \\
cyclist       & 1775  & 1225  & 856       & 1122     & 1022     & 1283  & 1438     & 279   & 2384       & 577         & 39      \\
car           & 13045 & 11773 & 4896      & 7692     & 12230    & 13322 & 8541     & 2955  & 14711      & 2475        & 7632    \\
truck         & 9053  & 2034  & 403       & 7715     & 2969     & 5258  & 4919     & 910   & 3297       & 1015        & 6775    \\
tram/bus      & 1062  & 1484  & 951       & 236      & 1359     & 1515  & 565      & 466   & 2276       & 79          & 191     \\
tricycle      & 105   & 69    & 69        & 89       & 16       & 100   & 74       & 0     & 152        & 1           & 21      \\ \bottomrule
\end{tabular}%
}
\caption{Number of objects for each domain and each class in CLAD-C test set}
\label{tab:test_counts}
\end{table*}

\begin{table*}[]
\centering
\resizebox{0.8\textwidth}{!}{%
\begin{tabular}{@{}lll@{}}
\toprule
date       & period  & \#objects \\ \midrule
11/11/2019 & Daytime & 5157      \\
           & Night   & 1154      \\
17/11/2019 & Daytime & 6742      \\
           & Night   & 2560      \\
20/11/2019 & Daytime & 4517      \\
21/11/2019 & Night   & 2119      \\ \bottomrule
\multicolumn{3}{c}{\textbf{train}} \\ 
\end{tabular} \qquad
\begin{tabular}{@{}lll@{}}
\toprule
date       & period  & \#objects \\ \midrule
12/08/2018 & Daytime & 72        \\
15/10/2018 & Daytime & 973       \\
           & Night   & 28        \\
18/08/2018 & Daytime & 25        \\
15/01/2019 & Daytime & 2287      \\
29/06/2019 & Night   & 336       \\
11/10/2019 & Daytime & 31        \\
22/10/2019 & Night   & 1666      \\
28/10/2019 & Night   & 887       \\ \bottomrule
\multicolumn{3}{c}{\textbf{validation}} \\ 
\end{tabular} \qquad
\begin{tabular}{@{}lll@{}}
\toprule
date       & period  & \#objects \\ \midrule
15/10/2018 & Daytime & 17326     \\
           & Night   & 249       \\
05/01/2019 & Daytime & 199       \\
15/01/2019 & Daytime & 62        \\
19/06/2019 & Daytime & 115       \\
21/06/2019 & Daytime & 9301      \\
           & Night   & 523       \\
22/10/2019 & Night   & 6749      \\
28/10/2019 & Night   & 9177      \\
11/11/2019 & Daytime & 65        \\
           & Night   & 3149      \\ \bottomrule
\multicolumn{3}{c}{\textbf{test}} \\ 
\end{tabular}
}
\caption{Counts per day and period for train, validation and test set of CLAD-C. Note that only in the training set the order matters. Both the validation and test set are only used in their entirety to validate how well a model has learned.}
\label{tab:day_counts}
\end{table*}

\begin{table*}[]
\centering
\resizebox{0.6\textwidth}{!}{%
\begin{tabular}{@{}cccccccc@{}}
\toprule
                            &        & Pedestrian & Cyclist & Car   & Truck & Tram/bus & Tricycle \\ \midrule
\multirow{4}{*}{Train}      & Task 1 & 4397       & 5885    & 21156 & 3870  & 1528     & 202      \\
                            & Task 2 & 30         & 11      & 5161  & 4689  & 281      & 1        \\
                            & Task 3 & 628        & 508     & 6224  & 1493  & 356      & 4        \\
                            & Task 4 & 84         & 285     & 1892  & 1234  & 154      & 5        \\ \midrule
\multirow{4}{*}{Validation} & Task 1 & 504        & 663     & 2300  & 427   & 153      & 25       \\
                            & Task 2 & 1          & 4       & 584   & 518   & 21       & 0        \\
                            & Task 3 & 81         & 61      & 652   & 182   & 46       & 0        \\
                            & Task 4 & 16         & 46      & 232   & 137   & 25       & 2        \\ \midrule
\multirow{4}{*}{Test}       & Task 1 & 1298       & 2104    & 9820  & 4262  & 668      & 64       \\
                            & Task 2 & 569        & 110     & 8801  & 8869  & 290      & 25       \\
                            & Task 3 & 3377       & 1169    & 13701 & 2087  & 1125     & 63       \\
                            & Task 4 & 776        & 396     & 7645  & 1897  & 755      & 10       \\ \bottomrule
\end{tabular}%
}
\caption{Number of objects in each task in the three split of CLAD-D.}
\label{tab:CLAD-D_object_counts}
\end{table*}

\end{document}